\documentclass[10pt,twocolumn,letterpaper]{article}

\usepackage{iccv}
\usepackage{times}
\usepackage{epsfig}
\usepackage{graphicx}
\usepackage{amsmath}
\usepackage{amssymb}
\usepackage{subfig}
\usepackage{bm}

\makeatletter\renewcommand\paragraph{\@startsection{paragraph}{4}{\z@}
  {.5em \@plus1ex \@minus.2ex}{-.5em}{\normalfont\normalsize\bfseries}}\makeatother

\def\eg{\textit{e.g.}}
\def\ie{\textit{i.e.}}

\newcommand{\model}{PAL}%
\newcommand{\modelFull}{Prototype-centered Attentive Learning}%

\usepackage[pagebackref=true,breaklinks=true,letterpaper=true,colorlinks,bookmarks=false]{hyperref}

\iccvfinalcopy % *** Uncomment this line for the final submission

\def\iccvPaperID{} % *** Enter the ICCV Paper ID here

\ificcvfinal\fi

\begin{document}

%%%%%%%%% TITLE
\title{Few-shot Action Recognition with Prototype-centered Attentive Learning}

\author{Xiatian Zhu\textsuperscript{1} \quad
	Antoine Toisoul\textsuperscript{1} \quad
	Juan-Manuel Pérez-Rúa\textsuperscript{1} \quad
	Li Zhang\textsuperscript{2}\thanks{Work done while at Samsung AI Center Cambridge} \quad \\
	Brais Martinez\textsuperscript{1} \quad
	Tao Xiang\textsuperscript{3}$^*$
	\\ [0.1cm]
	\textsuperscript{1} Samsung AI Center Cambridge, UK
	\quad
	\textsuperscript{2} Fudan University
	\quad
	\textsuperscript{3} University of Surrey
}

\maketitle
% Remove page # from the first page of camera-ready.
\ificcvfinal\thispagestyle{empty}\fi

%%%%%%%%% ABSTRACT
\begin{abstract}
Few-shot action recognition aims to recognize new action classes with few training samples. Most existing methods adopt a meta-learning approach with episodic training. In each episode, the few samples in a meta-training task are split into support and query sets. The former is used to build a classifier, which is then evaluated on the latter using a query-centered loss for model updating. There are however two major limitations: lack of data efficiency due to the query-centered only loss design and inability to deal with the support set outlying samples and inter-class distribution overlapping problems. 
In this paper, we overcome both limitations 
by proposing a new \modelFull{} (\model{}) model composed of two novel components. First, a prototype-centered contrastive learning loss is introduced to complement the conventional query-centered learning objective, in order to make full use of the limited training samples in each episode. 
Second,  \model{} further integrates a hybrid attentive learning mechanism that can minimize the negative impacts of outliers and promote class separation. Extensive experiments on four standard few-shot action benchmarks show that our method clearly outperforms  previous state-of-the-art methods, with the improvement particularly significant ($>10\%$) on the most challenging fine-grained action recognition benchmark.

\end{abstract}
%%%%%%%%% BODY TEXT
\section{Introduction}
The universal adoption of smartphones has led to a large quantity of videos being produced and shared on social media platforms, which are in urgent need of automated analysis. As a result, video action recognition \cite{carreira2017quo,lin2019tsm,Tran2015C3D,S3D_G_ECCV18} has been studied intensively with a recent focus on fine-grained actions \cite{ssv1}. Most recent action recognition models  employ deep convolutional neural networks (CNNs) which are known to be data hungry as they require the collection of a large number (at least hundreds) of annotated samples for each action class to be trained. However, collecting and annotating such a large amount of data is expensive, tedious and sometimes even infeasible for some rare fine-grained action classes. 
This is the reason why few-shot action recognition has started to receive increasing interest  \cite{cao2020few,zhang2020few,zhu2018compound} as it aims at constructing a video action classifier with few training samples (\eg, 1-5) per class. 

\begin{figure}
    \centering
    \includegraphics[width=0.6\linewidth]{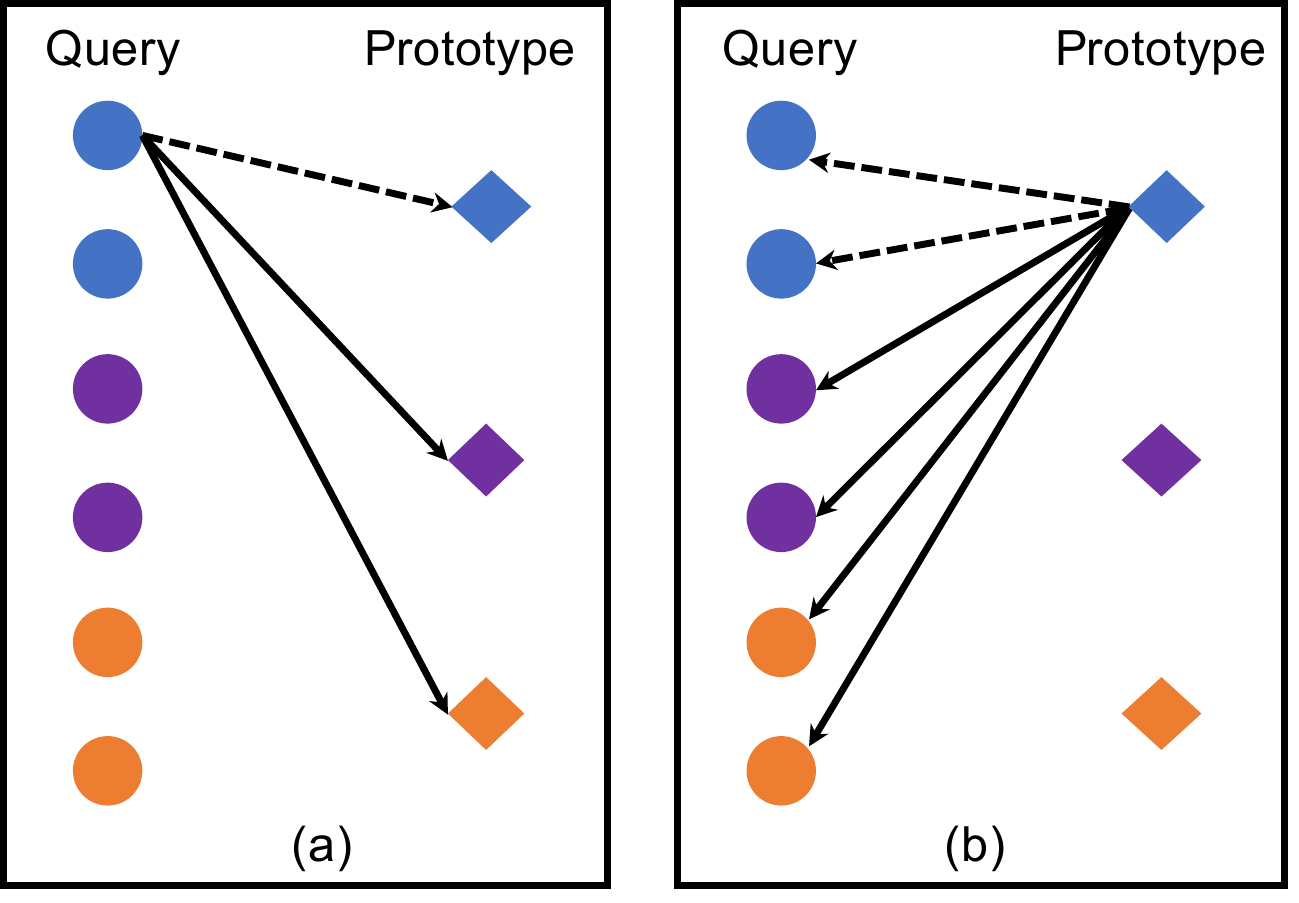}
    \caption{Illustration of (a) conventional query-centered learning vs. (b) our prototype-centered learning in a 3-way 2-shot case.
    Specifically, the former classifies a query sample against the space of prototype via computing pairwise similarity and softmax normalization. In contrast, the latter considers an opposite direction by comparing a prototype with all the query samples,
    pulling close those of the same class whilst pushing way the others. 
    Circle and diamond represent the query samples and per-class prototypes respectively.
    Each class is color coded.
    Dashed and solid arrows represent pairs from same and different classes.
    }
    \vspace{-0.6cm}
    \label{fig:concept}
\end{figure}

Few-shot action recognition is a special case of few-shot learning (FSL). Most FSL methods follow a meta-learning (or learning-to-learn) paradigm
\cite{snell2017prototypical,vinyals2016matching} characterized by episodic training which aims to learn a model or optimizer from a set of base/seen tasks, in order  to generalize well to  new tasks with few labeled training samples/shots. Specifically, a meta-training set containing abundant training samples per seen/base class is used to sample a large number of training episodes. In each episode, the training data is split into a support set with $N$ classes and $K$ samples per class to mimic the setting of target meta-test tasks, and a query set from the same $N$ classes. A classifier is built using the support set and then evaluated on each query set sample with a classification loss for model updating. Many state-of-the-art FSL methods  \cite{ye2020few,zhang2020deepemd} are based on the popular prototypical network (ProtoNet) \cite{snell2017prototypical} for its simplicity and effectiveness. With ProtoNet, the mean of each support set class is computed as a class prototype for classifying each query sample (see Figure \ref{fig:concept}(a)). 

There are however two major limitations in existing few-shot action recognition methods \cite{cao2020few,zhang2020few,zhu2018compound}. The first limitation is the lack of data efficiency due to the  query-centered learning objective adopted by existing methods. This is shown in Figure \ref{fig:concept}(a): with only $K$ samples for each of the $N$ classes in the support set, these $N \times K$ samples are first reduced to $N$ prototypes; a loss term (\eg, cross-entropy) is then computed for each query sample individually based on its distances to the prototypes, without any consideration on how the whole query set samples should be distributed. This query-centered learning objective thus  does not make full use of the limited training data in each episode.

\begin{figure}
    \centering
    \includegraphics[width=1\linewidth]{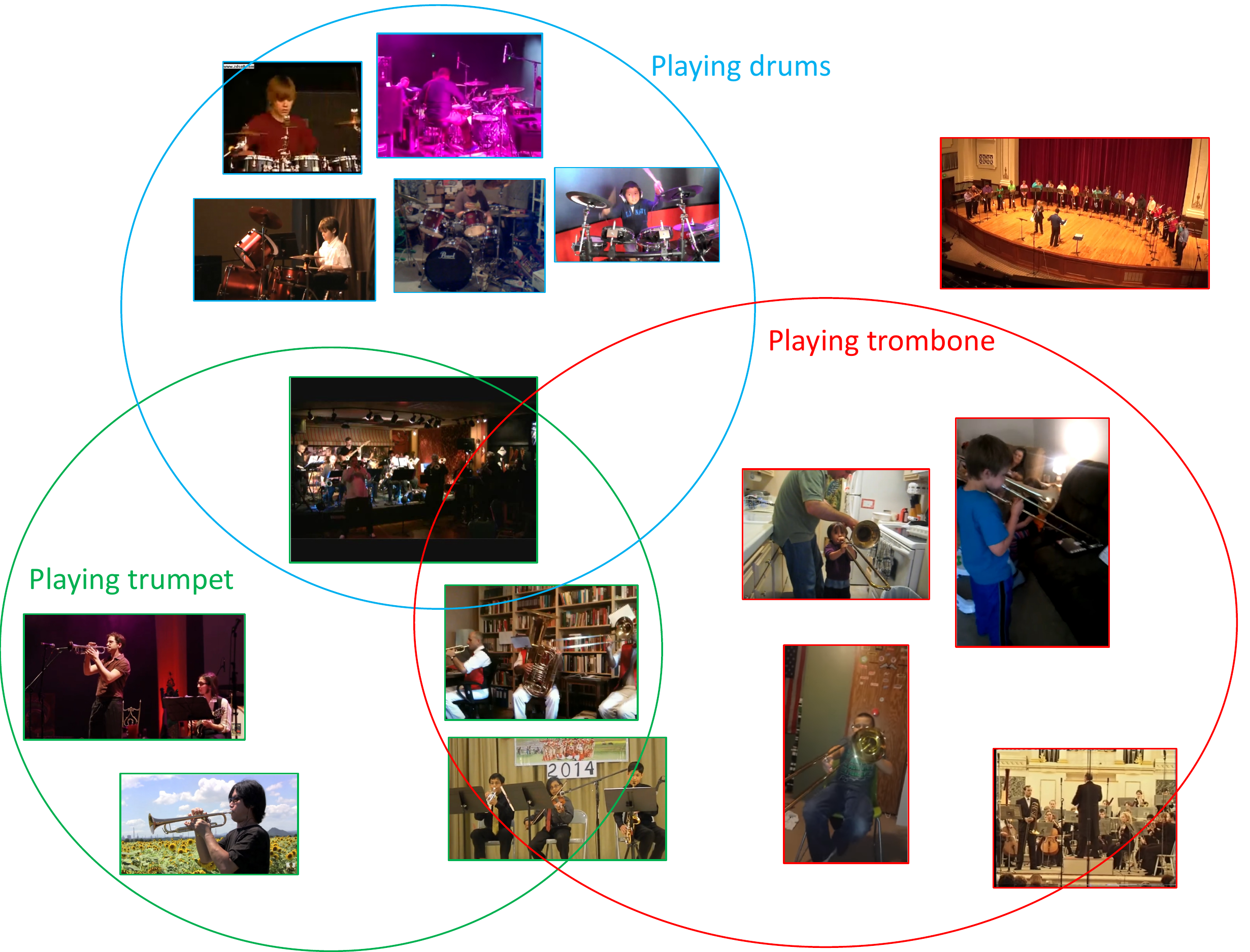}
    \caption{Illustration of the outlying video samples and inter-class overlapping problems in a support set containing three music-instrument-playing classes. There is a clear outlier in the trombone class due to its unusual view point. There is also plenty of class overlapping due to the intrinsic  visual similarity between the three classes, particularly when these instruments are played in groups. } 
\vspace{-0.6cm}
    \label{fig:outlier}
\end{figure}

The second limitation is their inability to address two fundamental challenges of FSL: outlying samples and inter-class distribution overlapping in the support set. These issues are in general more severe in video data than imagery due to high spatio-temporal redundancy.
As illustrated in Figure \ref{fig:outlier},  outliers are  caused by unusual viewpoint, background, occlusion etc. These are problematic for any learning task, but particularly for few-shot action recognition as with few samples per class, a single outlier can have an immense impact on the class distribution. The problem of inter-class overlapping is also common when the training samples of different classes have very similar background, or just being visually similar. This problem is especially acute for fine-grained action classes. Both problems are not addressed when a class is simply represented as class mean without considering both the intra-class relationships to identify outliers and inter-class relationships to avoid class overlapping.

To overcome the aforementioned limitations,
we propose a novel {\em \modelFull{}} (\model{}) model with two key components. First, a  {\em prototype-centered contrastive learning} loss is introduced. This loss is computed by comparing a given prototype 
against all the query-set samples for discriminative learning (see Figure \ref{fig:concept}(b)). This new learning objective is clearly complementary to conventional query-centered learning objective, and combining the two enables \model{} to make the full use of the limited training data and hence improve data efficiency during training.  Second, 
 a {\em hybrid attentive learning} component is developed 
consisting of self-attention on support-set samples
and cross-attention from query-set to support-set samples.
This design aims to mitigate the negative effect of
outlying samples and promote class separation. 
Importantly, it can be seamlessly integrated with the
query- and prototype-centered learning objectives
in a unified meta-learning pipeline. 

In this paper, we make the following {\bf contributions}:
(1) We propose a novel \modelFull{} (\model{})
for few-shot action recognition, specifically designed 
to address the data efficiency, outliers and class overlapping problems that existing methods suffer from.
(2) We introduce a novel prototype-centered contrastive learning objective to complement the existing query-centered objective, in order to 
make full use of the limited training data available in few shot learning.
(3) We further introduce a hybrid attentive learning strategy
which is dedicated for solving the under-studied outlier sample and class overlapping problems.
(4) Extensive experiments show that our model achieves new state-of-the-art results on four few-shot action datasets. Its performance is particularly compelling on the more challenging fine-grained benchmark, yielding around 10\% improvement.

%------------------------------------------------------------------------
\section{Related work}

\paragraph{Action recognition}
Previous efforts have been focused on training action recognition models on large-scale video datasets (\eg, the coarse-grained  Kinetics~\cite{carreira2017quo} and fine-grained Something-Something \cite{ssv1}).
Computationally, utilizing 2D CNNs~\cite{lin2019tsm,simonyan2014two,wang2016temporal} for action recognition is more efficient than using their 3D counterparts~\cite{carreira2017quo,lin2019tsm,Tran2015C3D,S3D_G_ECCV18}.
In particular, Temporal Segment Networks (TSN)~\cite{wang2016temporal} are one of the most popular action classification models. They work by extracting features from temporally sparsely sampled frames with a 2D CNN before applying an average pooling to obtain a video-level feature representation and a prediction.
We use TSN as the feature extractor in our few-shot action recognition model as we found that more complex feature extractors~\cite{carreira2017quo,lin2019tsm,Tran2015C3D,S3D_G_ECCV18} do not fare better performance but are computationally more demanding.

\paragraph{Few-shot action recognition}
When the research focus in action recognition shifted towards fine-grained action classes, the problem of collecting sufficient training samples per class started to become an obstacle.    Few-shot action recognition emerged as a potential solution to this problem ~\cite{cao2020few,zhang2020few,zhu2018compound}. CMN~\cite{zhu2018compound} employs a compound memory network to store the representation and classify videos by matching and ranking.
OTAM~\cite{cao2020few} measures the query-centered distance with respect to support set samples, by explicitly leveraging the temporal ordering information in the query video via ordered temporal alignment. 
ARN~\cite{zhang2020few} learns a query-centered similarity between the query and support video clips with a pipeline following RelationNet~\cite{sung2018learning}.
It exploits augmentation-guided spatial and temporal attention with auxiliary self-supervision training losses.
Concurrently introduced along with our work,
TRX~\cite{perrett2021temporal} leverages
sub-sequence level self-attention to form query 
specific prototypes at varying temporal scales.
This allows to mitigate the temporal misalignment problem in a partial matching manner.
%AmeFu-Net~\cite{fu2020depth} leverages depth information to assist computation of background and foreground, which encourage the model to capture more temporal-sensitive information. 
Despite their differences in model design, all deploy a meta-learning framework with a query-centered learning objective, thus being unable to make full use of the limited training data available in each episode. Furthermore,  none is designed to address the 
 the outlying sample and class overlapping problems, as our model does.
%in the few-shot action recognition context.

%\vspace{-0.3cm}
\paragraph{Few-shot learning}
Few-shot action recognition is a special case of few-shot learning (FSL). Most existing FSL models focus on recognizing static images. They can be roughly categorized into two groups including metric-based methods~\cite{doersch2020crosstransformers,perez2020incremental,snell2017prototypical,sung2018learning,vinyals2016matching,ye2020few,zhang2020deepemd} and optimization-based methods~\cite{finn2017model,ravi2016optimization}. All existing few-shot action recognition methods and our \model{} belong to the first group, where what is meta-learned is a feature embedding network for videos. Note that the idea of using 
self-attention for few-shot learning has previously been explored in~\cite{ye2020few}. However, the model in \cite{ye2020few} only applies self-attention to  class prototypes whereas, in contrast, our model applies it to the support set as well as query samples. This gives our model the capability of dealing with sample outliers while improving data efficiency. Following the practice of most recent FSL methods \cite{ye2020few,zhang2020deepemd}, we also incorporate pretraining of the feature embedding network into our model.

%\vspace{-0.3cm}
\paragraph{Self-attention}
Our attentive learning is based on self-attention across data instances, which has been first introduced in transformers  as a global self-attention module for machine translation tasks~\cite{vaswani2017attention}.
Non-local neural networks~\cite{wang2018non} applied the core self-attention block from transformers for context modeling and feature learning in computer vision tasks.
They learn an affinity map among all pixels in the image that allows the neural network to effectively increase the receptive field to the global context.
State-of-the-art performances have been shown in classification~\cite{dosovitskiy2020image}, self-supervised learning~\cite{chen2020generative}, semantic segmentation~\cite{fu2019dual,zhang2020dynamic}, object detection~\cite{carion2020end,yin2020disentangled,zhu2020deformable}
% , and few-shot visual recognition~\cite{doersch2020crosstransformers} 
by using such a transformer-based attention model.
Different from these works, in this paper, we propose a hybrid attentive learning mechanism that aims to exploit the support set \textit{self-attention} and query-to-support \textit{cross-attention} in a unified meta-learning manner.

\begin{figure*}
    \centering
    \includegraphics[width=1\linewidth]{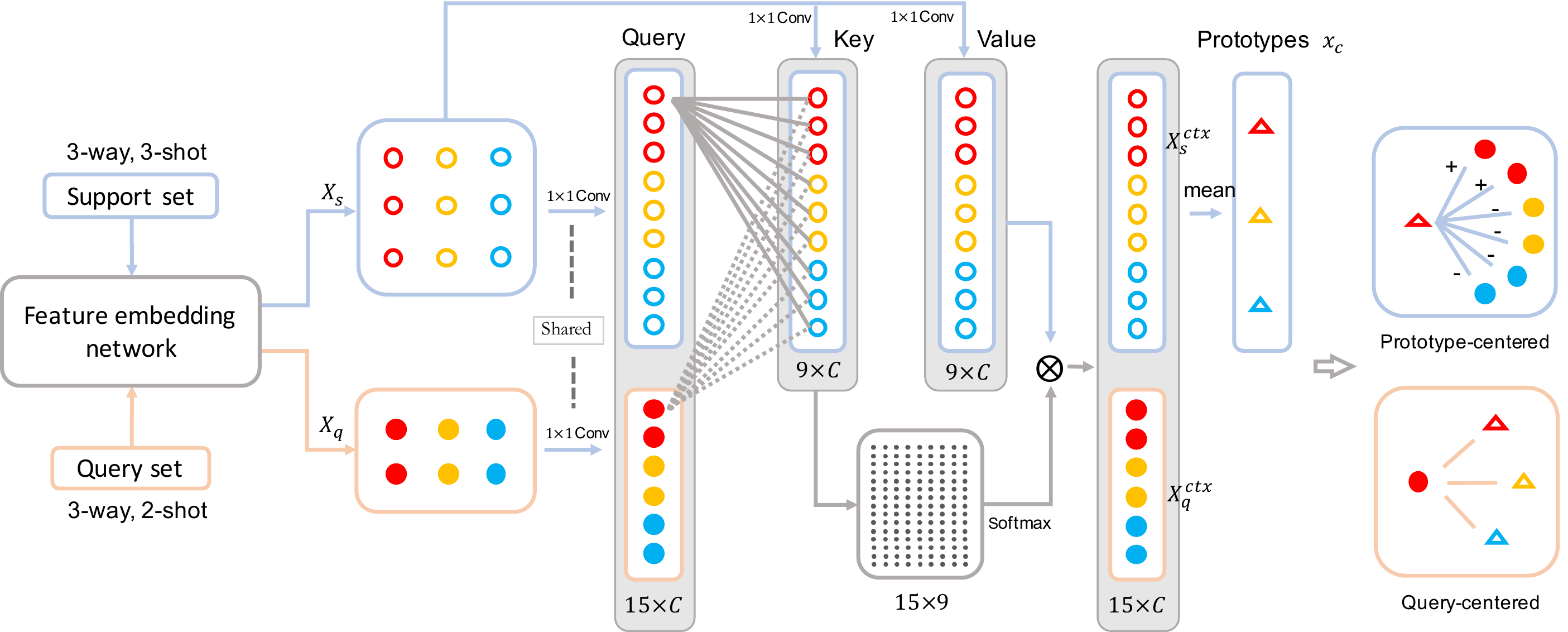}
    \caption{
    Schematic illustration of the proposed \modelFull{} (\model{}).
    %\model{} is designed to solve the challenging inter-class distribution overlap problem in few-shot action recognition.
    Given an episode of training video data including a support set (3-way 3-shot) and a query set (2 samples per class), a CNN feature embedding network is first used to extract their feature vectors $\textbf{X}_s$ and $\textbf{X}_q$. 
The feature vectors $\textbf{X}_s$ of all support samples are then selected to form  the \textit{Key} and \textit{Value} via respective linear transformation. 
    For {\em support-set self-attention} learning,
    the transformation of $\textbf{X}_s$ is set as the \textit{Query} and used to conduct attentive learning with \textit{Key} and \textit{Value}.
    For {\em query-to-support cross-attention learning},
    $\textbf{X}_q$ is instead used as the input of \textit{Query}.
    The two processes form a {\em hybrid attentive learning} strategy for solving the outlying sample problem, which outputs attentive support-set $\textbf{X}_s^\text{ctx}$ and query-set $\textbf{X}_q^\text{ctx}$ features.
 After obtaining the per-class prototypes $\textbf{x}_c$ from $\textbf{X}_s^\text{ctx}$ as feature mean, $\textbf{X}_q^\text{ctx}$ can be classified by computing pairwise similarity against $\textbf{x}_c$ and softmax normalization (\ie, query-centered learning).
 To further improve discriminative learning of limited training data, we additionally conduct a novel prototype-centered contrastive learning 
    in an opposite direction
    by comparing a given prototype with all the query-set samples and conducting discriminative contrastive learning.
    % {\bf \color{red}{TODO: Li is refining}}
    }
    \label{fig:pipeline}
\end{figure*}

\section{Method}

% \noindent {\bf Problem definition.}
\subsection{Problem definition}
\label{sec:problem}
We consider the standard few-shot video action classification problem definition \cite{cao2020few,zhu2018compound}.
Given a meta-test dataset $D_\text{test}$,
we sample a $N$-way $K$-shot classification task
to test a learned FSL action model $\bm{\theta}$.
To train the model $\bm{\theta}$ in a way that
it can perform well on those sampled classification tasks,
episodic training is adopted to meta-learn the model.
Concretely, a large number of 
$N$-way $K$-shot tasks are randomly sampled 
from a meta-training set $D_\text{train}$,
and then used to train the model in an episodic manner.
In each episode, we start by sampling $N$ classes 
from $D_\text{train}$ at random. Then we randomly draw labeled training samples from these classes to create a {\em support} set $\mathcal{S}$
and a {\em query} set $\mathcal{Q}$ consisting of 
$K$ and $Q$ samples per class, respectively.
Formally, the support and query sets are defined as:
\begin{align}
    \mathcal{S} = \{ (x_i, y_i) \; | \; y_i \in C \}_{i=1}^{NK},\\
    \mathcal{Q} = \{ (x_i, y_i) \; | \; y_i \in C \}_{i=1}^{NQ}.
\end{align}
Note, $\mathcal{S} \cap \mathcal{Q} = \phi$ are sample-wise non-overlapping.

Typically, episodic training is conducted in a two-loop manner \cite{snell2017prototypical}:
the support set is used in the inner loop to construct a classifier for the $N$ classes,
and the query set is then used in the outer loop to evaluate this classifier and update the model parameters $\bm{\theta}$.
It is noteworthy that as the objective is to obtain a learner able to recognize novel classes each with only a few labeled examples,
$D_\text{train}$ and $D_\text{test}$ are set to be disjoint
in the class space, \ie,
$D_\text{train} \cap D_\text{test} = \phi$.
Unlike the sparsely annotated meta-test classes, 
each meta-training class comes with abundant labeled training data that allows to sample sufficient episodes for meta-training.

%-------------------------------------------------------------------------
\subsection{Pretraining}
\label{sec:baseline}

As in all existing few-shot action classification models \cite{cao2020few,zhang2020few,zhu2018compound}, one of the key objectives is to learn a feature embedding network through meta learning, so that it can generalize to any unseen action class. The embedding networks adopted   
are based on those used by existing video action recognition model with TSN \cite{wang2016temporal} being the most popular choice.
For example, in the state-of-the-art OTAM~\cite{cao2020few} model,  a ResNet-50 based TSN model, pre-trained on ImageNet,
was meta-trained together with a time-warping based video distance. However, in most recent FSL methods for static image classification \cite{ye2020few,zhang2020deepemd}, pretraining the embedding network on the whole meta-training set before episodic training starts has become a must-have step. In this work, we also adopt such a pretraining step and show in our experiments that this step is vital (see Sec.~\ref{sec:exp}).

Specifically, we use as our feature embedding network
a TSN action model \cite{wang2016temporal}. It is then pretrained on the whole training set $D_\text{train}$ with a cosine similarity based cross-entropy loss. 
Given a video sample $x_i \in D_\text{train}$ with a varying length and redundant temporal information, we first 
sample video frames. We adopt the same sparse sampling strategy as introduced in \cite{wang2016temporal}: 
splitting the video into $T$ equal-sized segments
and randomly selecting one frame from each segment.
This way, each video can be represented using a fixed number 
of frames $x_i=\{f_{1}, f_{2}, \cdots, f_{T}\}$
where $f_{j}$ is a random frame from the $j$-th segment.
As sampled frames cover the majority of original time span, 
long-term temporal information is kept whereas the spatio-temporal redundant information is reduced significantly.

Next, each sampled video frame $f_j$ is individually encoded by a feature embedding network $h()$ and classified by a cosine-distance based classifier,
giving us frame-level classification score vectors
$\textbf{s}^f_j = [s^f_1, s^f_2, \cdots, s^f_Z]$
with $Z$ the total number of classes in $D_\text{train}$.
In particular, for each frame, $s^f_z = \cos (h(f_j), \; \textbf{w}_z) $ where $\cos()$ denotes the cosine similarity function and $\textbf{w}_z$ ($z \in \{1, \cdots, Z\}$) the classifier weights of each class. Then, these per-frame score vectors are averaged to get video-level scores :

\begin{align}
    \textbf{s}_i 
    & = \frac{1}{T}\sum_{j=1}^T \textbf{s}^f_j \\ \nonumber
& = [s_1, s_2, \cdots, s_Z]
\end{align}

Finally, a softmax is applied to the video-level scores to get video-level action probabilities : 

\begin{equation}
    p_z = \frac{\exp(s_z)}
    {\sum_{i=1}^Z \exp(s_i)} \text{ for } z \in \{1, \cdots, Z\}
    \label{eq:sm}
\end{equation}

During training, $\textbf{p}_i = [p_1, p_2, \cdots, p_Z]$ and its ground-truth class label $y_i$ are then employed to optimize the model using a cross-entropy loss:
\begin{equation}
    \mathcal{L}_{ce} = - \sum_{z=1}^Z \mathbb{I}(z==y_i) \log p_z,
    \label{eq:ce}
\end{equation}
where $\mathbb{I}()$ is an indicator function which returns 1 when the argument is true, 0 otherwise.

After the pretraining stage, the feature embedding network together with a cosine distance can be used directly for meta-test without going through the episodic training stage. Our experiments  show that this turns out to be a surprisingly strong baseline that even achieves better results than current state-of-the-art OTAM \cite{cao2020few} (see Table \ref{tab:sota}).
This verifies for the first time that a good feature embedding (or feature reuse) is also critical for few-shot action modeling -- a finding that has been reported in recent static image FSL works  \cite{wei2019Close,liu2020negative,tian2020rethinking}.
Nonetheless, this baseline is still limited for FSL
since it lacks a ``learning to learn'' or task adaptation capability to better deal with unseen
new tasks. 

To this end, we next introduce our {\em \modelFull{}} (\model) method.
The overview of \model{} is depicted in Figure \ref{fig:pipeline}.
\model{} is built upon the ProtoNet \cite{snell2017prototypical} but consists of two new components:
(1) Hybrid attentive learning including 
self-attention on support-set samples
and cross-attention from query-set samples to support-set samples
(Sec.~\ref{sec:hal}).
(2) Prototype-centered contrastive learning (Sec.~\ref{sec:pcc}).

%-------------------------------------------------------------------------
\subsection{Hybrid attentive learning}
\label{sec:hal}
Hybrid attentive learning (HAL) is designed to mitigate the inter-class ambiguity and the intra-class outlying sample problem by allowing each support set sample to examine all other samples in order to identify and fix both problems. 
It relies on a transformer self-attention module whose parameters are meta-learned together with those of the embedding network $h()$. Once learned, during meta-test, it is used to exploit task-specific contextual information for 
superior task adaptation.
% into both support-set and query-set samples. 
%
Concretely, given an episode 
we first extract a $d$-dimensional feature representation using a TSN $h()$.
The video-level feature vectors are obtained for
both support-set and query-set by average pooling along frames.

\paragraph{ Support set self-attention}
Formally, let $\textbf{X}_s  \in \mathbb{R}^{NK \times d}$ and $\textbf{X}_q \in \mathbb{R}^{NQ \times d}$ be the support-set and query-set feature matrix respectively. 
The input to HAL is in the triplet form of
(\texttt{Query}, \texttt{Key}, \texttt{Value}). 
To learn discriminative contextual information per episode,
the input is designed based on the support-set samples as:
\begin{align}
    \text{Query} = \textbf{X}_s \textbf{W}_Q, \;\;  
    \text{Key} = \textbf{X}_s \textbf{W}_K, \;\;
    \text{Value} = \textbf{X}_s \textbf{W}_V,
    \label{eq:qkv}
\end{align}
where $\textbf{W}_Q$/$\textbf{W}_K$/$\textbf{W}_V  \in \mathbb{R}^{d\times d_a}$
are learnable parameters 
(each represented by a fully connected layer) 
that project the TSN feature to a $d_a$-D latent space.
As \texttt{Query}, \texttt{Key} and \texttt{Value} share the same input source (\ie, support-set data),
{\em self-attention} can be formulated as:
\begin{equation}
    \textbf{X}_s^\text{ctx} = \textbf{X}_s + 
    \operatorname{softmax}(\frac{\textbf{X}_s \textbf{W}_Q (\textbf{X}_s \textbf{W}_K)^\top}{\sqrt{d_a}}) (\textbf{X}_s \textbf{W}_V),
    \label{eq:attn}
\end{equation}
where $\operatorname{softmax}()$ is a row-wise softmax function for attention normalization.
Residual learning is adopted for a more stable model convergence.
As written in Eq \eqref{eq:attn}, pairwise similarity 
defines the attentive scores between any two support-set samples,
and further used for weighted aggregation in the \texttt{Value} space.
The intuition is that, statistically sample pairs from the same class often enjoy more similarity than those with different classes except for very few outlier instances. As a result, this attentive learning reinforces the importance of class-sensitive information, 
subject to the context of current task's classes.
Consequently, the effect caused by class irrelevant information of outlier samples will be well controlled during feature transformation.
Finally, intra-class variation shrinks and 
inter-class ambiguity can be reduced accordingly. 
%
%The output $\textbf{X}_s^\text{cxt}$ is thus a task-adapted support-set representation.

\paragraph{ Query set cross-attention} For consistency, the query-set samples should be also contextualized
in a task-specific manner, because they will be classified into a label space formed by support-set samples.
% since their TSN features contain no task-specific contextual information.
To that end, we introduce a {\em cross-attention} process from query-set samples 
$\textbf{X}_q$ to support-set samples $\textbf{X}_s$ formulated as:
\begin{equation}
    \textbf{X}_q^\text{cxt} = \textbf{X}_q + 
    \operatorname{softmax}(\frac{\textbf{X}_q \textbf{W}_Q (\textbf{X}_s \textbf{W}_K)^\top}{\sqrt{d_a}}) (\textbf{X}_s \textbf{W}_V).
    \label{eq:xattn}
\end{equation}
By doing this, in the same spirit as support-set self-attention,
query-set samples are also enriched by contextual information. Note that each query set sample transformation is done independently from other query samples as the \texttt{Keys} and \texttt{Values} are provided by the support set only. Our model thus remains inductive as existing FSL action models. 

\paragraph{Learning objective}
To obtain the supervision for our HAL module, we adopt the popular prototype based objective loss
\cite{snell2017prototypical}. With the attentive support-set feature matrix $\textbf{X}_s^\text{ctx}$, we first form the prototype for each class $c$ as the mean feature:
\begin{equation}
    \textbf{w}_c = \frac{1}{K} \sum_{i = 1}^{NK} \mathbb{I}(y_i == c) \textbf{X}_s^\text{ctx}(i, :),
\end{equation}
where $y_i$ denotes the class label of $i$-th support-set sample. 
We then compute the cosine similarity between any query-set feature
$\textbf{X}_q^\text{ctx} (j,:)$ and all prototypes $\{\textbf{w}_c\}_{c=1}^N$ and obtain the classification probability vector $\tilde{\textbf{p}}_j = [\tilde{p}_1, \cdots, \tilde{p}_N] \in \mathbb{R}^N$ by softmax function (Eq \eqref{eq:sm}).
With the query-set class labels $y_j$, a cross-entropy loss 
can be derived over $N$ classes of the current episode as the meta-training objective:
\begin{equation}
    \mathcal{L}_{meta} = - \sum_{n=1}^N \mathbb{I}(n==y_j) \log \tilde{p}_n,
    \label{eq:meta_ce}
\end{equation}
Which will be used to update the parameters of both the HAL module and the TSN embedding network.

\subsection{Prototype-centered contrastive learning}
\label{sec:pcc}

The meta-loss in Eq. \eqref{eq:meta_ce} is still the conventional query-centered loss (see Figure \ref{fig:concept}(a)).
That is, only query-to-prototype discrimination
is considered, whilst the other way around is ignored.
We hypothesize that this design fails to make full use of 
already-limited training data in each episode and 
would lead to sub-optimal task adaptation.
%Consequently, it becomes less effective in resolving inter-class ambiguity underlying in new tasks.

To overcome this limitation, we further introduce 
a complementary prototype-centered contrastive learning. 
More specifically, for a prototype $\textbf{w}_c$,
we define the query-set samples $\mathcal{Q}_c$ from the same class as the positive matches
and all the others as the negative.
We then compute the cosine similarity between 
$\textbf{w}_c$ and all the query-set samples,
and devise the following prototype centered contrastive loss function:
\begin{equation}
    \mathcal{L}_{pcc} = \frac{1}{N} \sum_{c=1}^N 
    -\log \Big( \frac{\sum_{i \in \mathcal{Q}_c} \cos(\textbf{w}_c, \textbf{X}_q^\text{ctx} (i,:))}
    {\sum_{j \in \mathcal{Q}} \cos(\textbf{w}_c, \textbf{X}_q^\text{ctx} (j,:) )} \Big),
    \label{eq:pcc}
\end{equation}
where $N$ is the number of class prototypes.
This design attempts to pull positive query-set samples closer to their corresponding prototype (\ie, the anchor of our loss),
whilst pushing away the negative ones.
This helps to further reduce intra-class variation
while simultaneously better separate different classes.
Eq \eqref{eq:pcc} is designed similarly in spirit 
to previous supervised contrastive learning objective \cite{khosla2020supervised}
and neighborhood component analysis \cite{goldberger2004neighbourhood}.
However, our loss function fundamentally 
differs from them because (1) 
% it acts uniquely as a meta-learning loss;
unlike the existing supervised contrastive loss which is used for supervised learning, our loss is used in an episodic few-shot learning setting, aiming to learn task-agnostic models, \ie, meta-models.
and (2) it is specially designed for
posing prototype centered discrimination
in a FSL context with prototypes as anchors.

%-------------------------------------------------------------------------
\subsection{Objective loss, model training and inference}

The overall objective loss function of our \model{} model for meta-training is 
defined as:
\begin{equation}
    \mathcal{L} = \mathcal{L}_{meta} + \lambda \mathcal{L}_{pcc},
\end{equation}
where $\lambda$ is a weight hyper-parameter which 
we set to 1.

In summary, the proposed method is trained in two sequential stages:
{\em First}, the baseline TSN is trained on meta-training set $D_\text{train}$
in a standard supervised learning way (Sec.~\ref{sec:baseline});
{\em Second}, the feature embedding network (TSN)  and our 
\model{} model are further optimized end-to-end
in the episodic training stage.
During test, the model is fixed and can be used similarly as in training 
by forming prototypes with labeled support-set samples in order to classify each unlabeled query-set samples on a meta-test set $D_\text{test}$
with the cosine similarity based classifier model.
% in a nearest neighbor fashion.

\begin{table*}[h]%[!h]
 	\footnotesize
	\centering
	\setlength{\tabcolsep}{0.3cm} 
    \renewcommand{\arraystretch}{1.2}
	\begin{tabular}{l||cc|cc|cc|cc}
		\hline
		%& 
		& \multicolumn{2}{c|}{\bf Kinetics-100}
		& \multicolumn{2}{c|}{\bf Sth-Sth-100}
		& \multicolumn{2}{c|}{\bf HMDB51}
		& \multicolumn{2}{c}{\bf UCF101}
		\\
		\bf Method 
	%	& Modality
		& \bf 1-shot & \bf 5-shot 
		& \bf 1-shot & \bf 5-shot 
		& \bf 1-shot & \bf 5-shot 
		& \bf 1-shot & \bf 5-shot 
		\\
		\hline
		Matching Net$^\dagger$ \cite{vinyals2016matching}
	%	& RGB
		& 53.3 & 74.6 
% 		& - & - 
		& - & -
		& - & - 
		& - & - 
		\\
		MAML$^\dagger$ \cite{finn2017model}
		%& RGB
		& 54.2 & 75.3 
		& - & -
		& - & - 
		& - & - 
		\\
		ProtoNet++$^*$ \cite{snell2017prototypical}
	%	& RGB
		& 64.5 & 77.9
		& 33.6 & 43.0
		& - & - 
		& - & - 
		\\
		\hline
		TARN$^*$ \cite{bishay2019tarn}
	%	& RGB
		& 64.8 & 78.5
		& - & - 
		& - & - 
		& - & - 
		\\
		TRN++$^*$ \cite{zhou2018temporal} 
	%	& RGB
		& 68.4 & 82.0
		& 38.6 & 48.9
		& - & - 
		& - & - 
		\\
		\hline
		CMN \cite{zhu2018compound}
	%	& RGB
		& 60.5 & 78.9 
		& - & -
		& - & - 
		& - & - 
		\\
		CMN++$^*$ \cite{zhu2018compound}
		%& RGB
		& 65.4 & 78.8 
		& 34.4 & 43.8
		& - & - 
		& - & - 
		\\
		OTAM \cite{cao2020few}
	%	& RGB
		& 73.0 & 85.8 
		& 42.8 & 52.3
		& - & - 
		& - & - 
		\\
		ARN \cite{zhang2020few}
	%	& RGB
		& 63.7 & 82.4
		& - & -
		& 45.5 & 60.6
		& 66.3 & 83.1
		\\
		\hline
		FEAT \cite{ye2020few}
		& 74.0 & 86.5 & 45.3 & 61.2
		& 60.4 & 75.2
		& 83.9 & 94.5
		\\
		\bf \model{} (Ours)
	%	& RGB
		&\bf 74.2 & {\bf 87.1}
		&\bf 46.4 & \bf 62.6
		& {\bf 60.9} & {\bf 75.8}
		& {\bf 85.3} & {\bf 95.2}
		\\
		\hline
	\end{tabular}
	\vskip 0.1cm
	\caption{
	Few-shot action classification results.
	$^*$: Results from \cite{cao2020few};
	$^\dagger$: Results from \cite{zhu2018compound}; ProtoNet++: cosine-distance is used; TRN++: same as ProtoNet++ but with TRN feature embedding instead of TSN; CMN++: also with TRN embedding. 
	}
	\label{tab:sota}
\end{table*}

%------------------------------------------------------------------------
\section{Experiments}
\label{sec:exp}

\paragraph{\bf Datasets}
We used four few-shot action benchmarks in our evaluations.
(1) {\em Kinetics-100} is a 100-class subset of
Kinetics-400 \cite{Kinetics}, and was introduced firstly in \cite{zhu2018compound} for few-shot action classification. We followed the same protocol: 
64/12/24 classes and 6389/1199/2395 videos for meta-training, meta-validation and meta-testing respectively.
Whilst Kinetics is one of the most commonly evaluated datasets, visual appearance and background encapsulate
most class-related information rather than motion patterns
\cite{sevilla2019only}.
With less need for temporal modeling and involving coarse-grained action classes,
it presents a relatively easy action classification task.
We hence further evaluated on (2) {\em Sth-Sth-100} created in a similar way based on the Something-Something-V2 dataset \cite{ssv1}. For this dataset we adopted the same protocol
as \cite{cao2020few} where 64/12/24 classes and 66939/1925/2854 videos are included for train/val/test respectively. 
By considering fine-grained actions involving human object interactions with subtle differences between different classes, it presents a significantly more challenging action recognition task. 
To compare with very recent models \cite{zhang2020few},
we also tested two more YouTube video datasets under the same setting.
(3) {\em HMDB-51} \cite{kuehne2011hmdb} contains 51 action classes with 6,849 videos.
We used 31/10/10 action classes with 4280/1194/1292 videos for train/val/test, respectively.
(4) {\em UCF101} \cite{soomro2012ucf101} consists of 
101 action categories with 13,320 video clips. 
We took 70/10/21 classes with 9154/1421/2745 videos for train/val/test, respectively.

\vspace{-0.2cm}
\paragraph{\bf Evaluation metrics} We adopted the standard 5-way 1/5-shot FSL evaluation setting \cite{cao2020few,zhu2018compound}. We randomly sampled 150,000 episodes from the meta-test set and reported the mean classification accuracy.

\vspace{-0.2cm}
\paragraph{\bf Implementation details}
We used an ImageNet pretrained ResNet-50 \cite{he2016deep} as the backbone network. We replaced the original fully-connected layer with a new layer that performs cosine similarity based classification.
For the first training stage,
we optimized the TSN feature embedding \cite{wang2016temporal}
(Sec. \ref{sec:baseline}) with SGD, 
with a starting learning rate at 0.001 and decaying every 30 epochs by 0.1 and a total of 70 epochs.
For the second stage, we conducted meta-training 
of both the TSN feature backbone and our \model{} model
end-to-end. On Sth-Sth-100, from the initial learning rate at 0.0001 we trained a total of 35 epochs with decaying epochs at 15 and 30 and each epoch consists of 200 episodes. 
For the other datasets, we found that training 10 epochs sufficed, with decaying points at  
5, 7 and 9.
For both stages, we used a cosine based classifier model.
We set the dimension of feature and latent space to $d=d_a=2024$ (Eq. \ref{eq:qkv}).
During training, we resized each video frame to 
the size of $256 \times 256$ from which a random $224 \times 224$ region was cropped to form the input. 
For three coarse-grained datasets, we applied random horizontal flip during training. However, many classes in the fine-grained Sth-Sth-100 are direction-sensitive (\eg, pulling something from left to right and pulling something from right to left), horizontal flip was therefore not applied.
During test, we used the center crop only to get the prediction of test data and model performance.

\begin{figure*}
    \centering
    \subfloat[Before PAL]{%
       \includegraphics[width=0.49\textwidth, page=1]{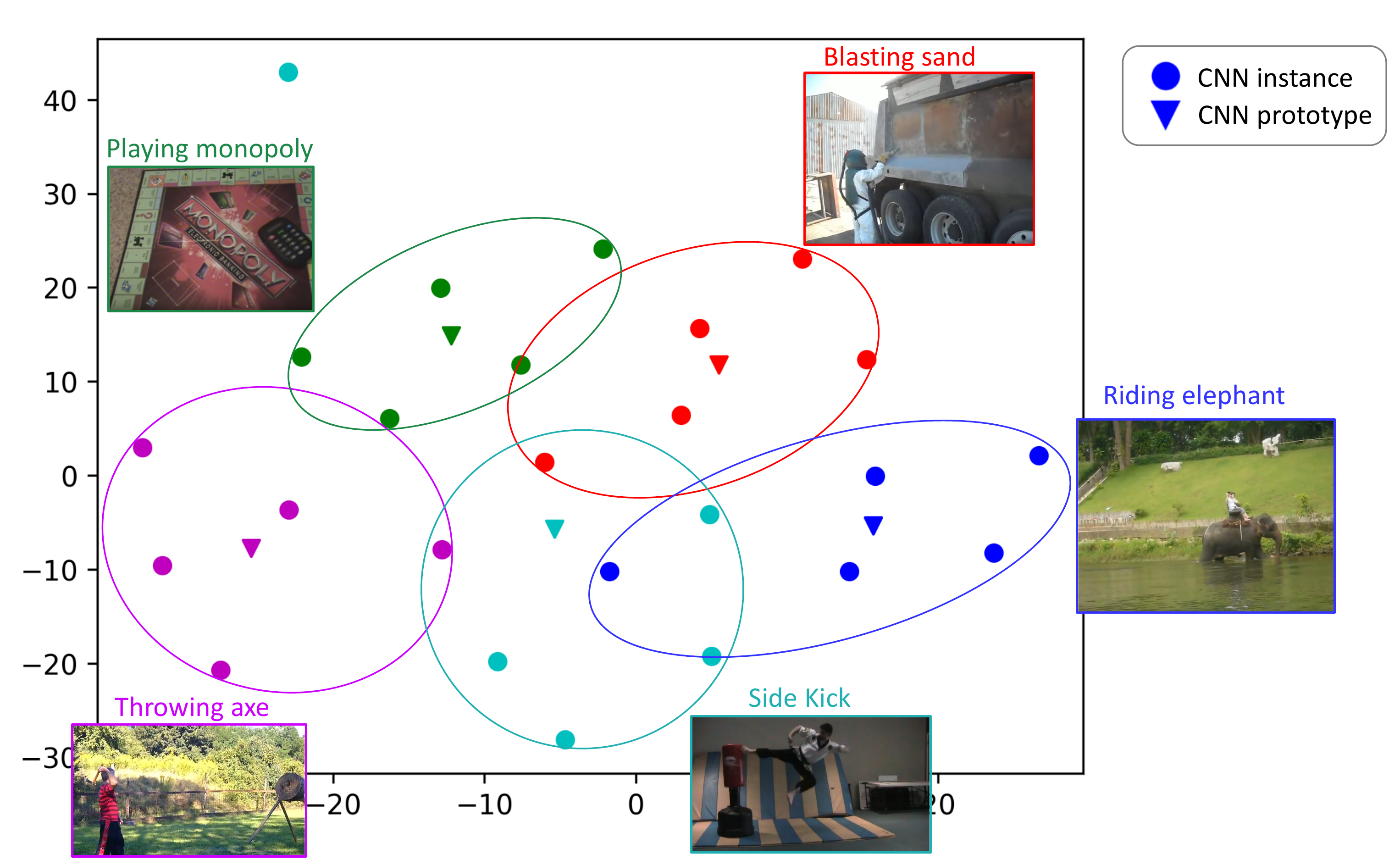}}\hspace{0.5em}
    \subfloat[After PAL]{%
       \includegraphics[width=0.49\textwidth, page=2]{figs/visualizations.pdf}}
    \caption{
    t-SNE feature projection of support samples
    and per-class prototypes \textbf{(a)} before and \textbf{(b)} after the proposed \model{}
    in a 5-way 5-shot task on Kinetics-100.
    Circle indicates the variation and boundary of each class. Each class is color coded. 
    As can be seen, the boundary of each class becomes tighter and the overlap between different classes is clearly reduced.
    }
    \label{fig:vis}
\end{figure*}

\paragraph{\bf Competitors}
Three groups of methods are compared in performance evaluation:
(1) Classical FSL models originally proposed for image classification
including Matching Net \cite{vinyals2016matching}, MAML \cite{finn2017model} and ProtoNet \cite{snell2017prototypical}.
All these FSL methods used the same TSN feature embedding for fair comparison.
(2) Stronger action recognition models including 
TRN \cite{zhou2018temporal} and TARN \cite{bishay2019tarn}.
(3) State-of-the-art FSL action models including 
CMN \cite{zhu2018compound},
OTAM \cite{cao2020few}, ARN \cite{zhang2020few},
and FEAT \cite{ye2020few}.

%------------------------------------------------------------------------
\subsection{Main results}

\paragraph{\bf Comparison to state-of-the-art}
The comparative results are reported in Table \ref{tab:sota}. 
We make the following observations:
{\bf(1)} In comparison to all classical FSL methods, 
the proposed \model{} is clearly superior under both settings
and on all datasets.
{\bf(2)} Whilst stronger action recognition models (\eg, TRN++ and TARN) improve the results,
they still lag behind our model by a large margin.
{\bf(3)} Interestingly, the first FSL action model CMN is shown to
be inferior to both TRN++ and TARN, suggesting that its memory network's merit is less critical than better temporal structure modeling. Directly combining CMN with TRN helps little (CMN++).
{\bf(4)} By taking a pairwise temporal alignment strategy,
the state-of-the-art OTAM further improves the performance. 
Nonetheless, it is still outperformed by our \model{}
particularly on the more challenging Sth-Sth-100 dataset (10.3\% improvement under 5-shot).
This is not surprising because OTAM's temporal warping
is functionally sensitive to outlier (less consistent) 
support-set videos as typically
encountered in fine-grained actions with human-object interactions.
Besides, as compared with our model, OTAM is inferior in leveraging few shots per class during meta-learning due to the lack of prototype-centered discrimination.
On the easier dataset Kinetics-100 where class overlapping is less a problem, our \model{} model
remains superior to OTAM by a smaller margin,
indicating that tackling outlying samples
is consistently a more effective strategy.
{\bf(5)} 
Similar to \model{},
FEAT \cite{ye2020few} also 
exploits a transformer \cite{vaswani2017attention} based self-attention for few-shot image classification.
However, unlike our model, 
it directly adapts the class prototypes (\ie, feature mean)
using self-attention alone; it is thus unable to deal with outlier samples in support set.
Moreover, there is no cross-attention and only conventional query-centered learning is performed
during meta-training.
As a result, our \model{} is consistently 
superior to FEAT.
{\bf(6)} 
On the two smallest datasets HMDB51 and UCF, the proposed method also sets a new state-of-the-art, due to its superior ability to deal with sparse training samples.

\paragraph{\bf Why \model{} works? }
As described in Introduction and Figure \ref{fig:concept},
our \model{} is specially designed to overcome the class overlap challenge caused by inter-class boundary ambiguity and outlier support-set samples intrinsic to new tasks.
To understand the internal mechanism of our model, we visualized the change of support samples' feature representations and per-class prototypes in a new 5-way 5-shot task. 
Concretely, this contrasted the feature distributions
with and without our \model{} model to reveal how the feature space is improved.
It is evident in Figure \ref{fig:vis} that 
\model{} does improve the separation of different classes'
decision boundary by posing two effects:
(1) reducing intra-class variation by mitigating the distracting effect of outlier samples in forming class decision boundary,
and 
(2) minimizing inter-class overlap by conducting query-centered and prototype-centered discriminative learning
concurrently.

%------------------------------------------------------------------------
\subsection{Ablation study}
We conducted ablation study of our method
on the most challenging fine-grained video action dataset Sth-Sth-100 \cite{ssv1} with more subtle inter-class differences.

\paragraph{\bf Importance of pretraining}
In our model, the feature embedding TSN model is pretrained on the whole training set before the episodic training stage. 
Instead, the current state-of-the-art OTAM \cite{cao2020few}
skipped this pretraining step.
Table \ref{tab:baseline} shows surprisingly that this pretraining step is vital. 
Our PAL model only with pretraining is already more effective than OTAM.
This suggests for the first time that a strong feature embedding is important for action classification in a FSL context, confirming the similar conclusion drawn in the image counterparts \cite{wei2019Close,liu2020negative,tian2020rethinking}.
To evaluate how our pretraining can benefit existing few-shot action models,
we integrated it in our own OTAM implementation due to the absence of officially released code. The results show that once the feature embedding is well pretrained, OTAM's time warping becomes insignificant,
even introducing a negative impact in 5-shot case.

\begin{table}[th]%[!h]
 	\footnotesize
	\centering
	\setlength{\tabcolsep}{0.5cm} 
    \renewcommand{\arraystretch}{1.0}
	\begin{tabular}{l||cc}
		\hline
		& \multicolumn{2}{c}{\bf Sth-Sth-100}
		\\
		\bf Method 
		& \bf 1-shot & \bf 5-shot 
		\\
		\hline \hline
		OTAM \cite{cao2020few}
		& 42.8 & 52.3 \\
		\hline
        Pretrained PAL
        & 42.7 &\bf 59.6
        \\
        \hline \hline
        Pretraining PAL + OTAM$^*$
        & \bf 43.1 & 58.4 \\
        \hline
	\end{tabular}
	\vskip 0.1cm
	\caption{
	Importance of pretraining.
	$^*$: Our implementation.
	}
	\label{tab:baseline}
\end{table}

\paragraph{\bf Contributions of model components}
To understand the benefits of the two components in our \model{},
namely Hybrid Attentive Learning (HAL) and
Prototype-centered Contrastive Learning (PCL),
we conducted a detailed component analysis 
by comparing the performances with and without them.
From Table \ref{tab:ablation} the following observations can be made.
(1) Our model pretraining sets out with strong performance.
(2) Each component is useful and the two components are complementary to each other in improving the classification accuracy. 
Concretely, the two components jointly improve the accuracy by 3.7\% and 3.0\%
over pretraining in the 1-shot and 5-shot settings, respectively.

\begin{table}[th]%[!h]
 	\footnotesize
	\centering
	\setlength{\tabcolsep}{0.8cm} 
    \renewcommand{\arraystretch}{1.2}
	\begin{tabular}{l|cc}
		\hline 
		& \multicolumn{2}{c}{\bf Sth-Sth-100}
		\\
		\bf Method 
		& \bf 1-shot & \bf 5-shot 
		\\
		\hline \hline
        \em Pretrained PAL
        & 42.7 & 59.6
        \\
        \hline \hline
	    HAL
		& 45.3 & 62.1
		\\
		\hline
		PCL
		& 44.5 & 61.4
		\\
		\hline
		HAL+PCL
		&\bf 46.4 & \bf 62.6
		\\
		\hline
	\end{tabular}
	\vskip 0.1cm
	\caption{
	Model component analysis.
    Two main components of our \model{} were evaluated.
	{\bf HAL}: Hybrid Attentive Learning;
	{\bf PCL}: Prototype-centered Contrastive Learning.
	}
	\label{tab:ablation}
\end{table}

%------------------------------------------------------------------------
\section{Conclusion}

In this work we have proposed a novel \modelFull{} (\model{}) method for few-shot action recognition.
It is designed specifically to address the data efficiency and inability to deal with outliers and class-overlapping problems of existing methods.
To that end, two complementary components are developed,
namely prototype-centered contrastive learning that allows to make better use of few shots per class,
and hybrid attention learning that aims to mitigate the negative effect of outlier support samples as well as class overlapping.
They can be integrated in a single framework
and trained end-to-end to maximize their complementarity.
%Moreover, we also contribute a strong baseline which we believe can offer a strong competitor and model initialization for future investigation.
Extensive experiments validate that the proposed \model{} 
yields new state-of-the-art performances on four action benchmarks, with the improvement on the more challenging fine-grained action recognition benchmark being the most compelling.

{\small
\bibliographystyle{ieee_fullname}
\bibliography{fsl_ref}
}

\end{document}